# Iterative Closest Labeled Point for Tactile Object Shape Recognition

Shan Luo[1], Wenxuan Mou[2], Kaspar Althoefer[1], and Hongbin Liu[1]

*Abstract*—Tactile data and kinesthetic cues are two important sensing sources in robot object recognition and are complementary to each other. In this paper, we propose a novel algorithm named Iterative Closest Labeled Point (iCLAP) to recognize objects using both tactile and kinesthetic information. The iCLAP first assigns different local tactile features with distinct label numbers. The label numbers of the tactile features together with their associated 3D positions form a 4D point cloud of the object. In this manner, the two sensing modalities are merged to form a synthesized perception of the touched object. To recognize an object, the partial 4D point cloud obtained from a number of touches iteratively matches with all the reference cloud models to identify the best fit. An extensive evaluation study with 20 real objects shows that our proposed iCLAP approach outperforms those using either of the separate sensing modalities, with a substantial recognition rate improvement of up to 18%.

## I. INTRODUCTION

The object shapes can be haptically assessed at two scales [1], i.e., local and global shapes. The former can be revealed by a single touch. It is analogous to the human cutaneous sense of touch, which is localized in the skin. The latter reflects the contribution of both cutaneous and kinesthetic inputs, e.g., contours that extend beyond the fingertip scale. In such case, intrinsic sensors, i.e., mechanoreceptors in joints, are also utilized to acquire the position and movement of the fingers/end-effectors in space to integrate local features to recognize the identity of the object. Here the kinesthetic cues are similar to human proprioception that refers to the awareness of the positions and movements of the body parts. Though the tactile features extracted from the local regions have been extensively studied in the recent robotics research [2]–[5], the use of kinesthetic cues to recognize the global object shapes is still largely unexplored.

In this paper, we propose a novel method named Iterative Closest Labeled Point (iCLAP) to integrate tactile and kinesthetic cues fundamentally to achieve a better object recognition performance. As illustrated in Fig. 1, the proposed iCLAP algorithm utilizes both appearance features extracted from tactile images and contact points in space. With only tactile readings, a Bag-of-Words (BoW) framework is first applied to the training dataset to form a dictionary of tactile features. By searching for the nearest word in the dictionary, each tactile feature is assigned a label number. With the positions of the object-tactile sensor interaction points in the 3D space, we can obtain a 4D data point cloud with the label number as the fourth dimension. The 4D point clouds of objects obtained during training are taken as reference models. The partial 4D point cloud obtained from testing iteratively matches with all the reference models and the identity of the best-fit reference model is assigned to the test object. To evaluate our proposed approach, 20 objects from either the lab environment or daily life were utilized in our experiments. The iCLAP algorithm was compared with two other methods using single sensing modalities in terms of classification of these objects: 1) BoW based object recognition with tactile data only [6]; 2) Iterative Closest Point (ICP) based method with kinesthetic data only [7]. Experiments show that our proposed iCLAP approach outperforms both methods. The experiments of classifying 20 real objects show that the classification performance is improved by up to 18% by using iCLAP compared to methods based on single sensing sources and a high average recognition rate of 85.36% can be achieved.

[1]Shan Luo, Kaspar Althoefer, Hongbin Liu are with the Centre for Robotics Research, Department of Informatics, King's College London, WC2R 2LS, UK (e-mail: shan.luo@kcl.ac.uk, k.althoefer@kcl.ac.uk, hongbin.liu@kcl.ac.uk).

[2]Wenxuan Mou is with School of Electronic Engineering and Computer Science, Queen Mary University of London, London E1 4NS, UK (e-mail: w.mou@qmul.ac.uk).

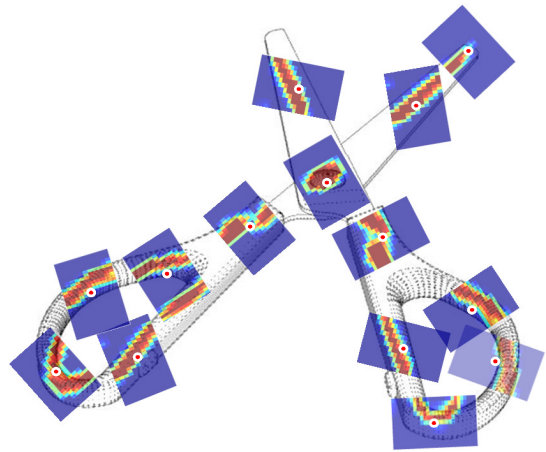

Fig. 1. The distribution of the collected data from a pair of scissors. At each collecting data point, the visualized tactile reading is present in a rectangle block and the 3D location of the tactile sensor where the tactile reading is gathered appears in a red round dot with white edges at the block center.

The remainder of this paper is organized as follows. The literature in the tactile shape recognition is reviewed in Section II. The iCLAP algorithm is introduced in detail in Section III, followed by an introduction of the data acquisition system in Section IV. The experiment results are provided and analyzed in Section V. In the last section the conclusions are drawn; possible applications and future research directions are also given.

## II. RELATED WORK

The current methods of haptic shape recognition can be divided into three categories according to the inputs: methods based on 1) the distributions of contact points; 2) the pressure patterns in tactile arrays; 3) both contact points and tactile patterns.

### A. Contact points based recognition

The methods based on contact points were widely used by early researchers due to the low resolution of tactile sensors and prevalence of single-contact force sensors. Allen *et al.* fit a super-quadric surface to sparse finger-object contact points and the parameters of the recovered super-quadrics were used for haptic object recognition [8]. In a similar manner, relying on the locations of the contact points and hand pose configurations, a polyhedral model was derived to recover the shape of an unknown object in [9]. More recently, Pezzementi *et al.* proposed a method to mosaic tactile measurements to construct an object representation [10]. Meier *et al.* [7] applied Kalman filters to generate 3D representations of unknown objects from contact point clouds collected by tactile sensors and classified the objects with the ICP algorithm. In [11], contact point clouds are combined with voxel representations to model the object shapes. By utilizing these methods, arbitrary contact shapes can be retrieved, however, it can be time consuming when investigating a large object surface as excessive contacts are required.

### B. Tactile patterns based recognition

Thanks to the recent development of tactile array sensors [12], the methods based on tactile patterns become popular in recent years. To decode the local contact shapes, various methods have been proposed. In [13], [14], the covariance of the pressure patterns were utilized to distinguish geometric shapes and estimate the poses of objects. Tactile patterns can also be treated as sparse images, and thus image features can be extracted to represent the data. Schneider *et al.* [6] first adapted the Bag-of-Words framework in the tactile object recognition and took tactile images as features directly. In [2] and [15], various feature descriptors from computer vision were studied in tactile image processing. Using a high resolution tactile imager named GelSight, Li *et al.* [16] employed binary descriptors to match features extracted from tactile readings and created tactile maps of objects. In our previous work [17], [18], a new Tactile-SIFT descriptor was proposed based on the Scale Invariant Feature Transform (SIFT) descriptor [19]. In addition to the above hand-drafted features, in [20] unsupervised hierarchical feature learning was applied to extract features from raw tactile data for grasping and object recognition tasks. In terms of recognizing the global object shape using tactile patterns, however, a limited number of approaches are available. One popular method is to generate a codebook of tactile features and use it to classify objects [4], [5], [14], [16]; a particular paradigm is the Bag-of-Words model. In this framework, only local tactile features are taken to generate a feature occurrence histogram to represent the object whereas in this type of methods the three-dimensional distribution information is not incorporated.

### C. Object recognition based on both sensing modalities

For humans, the sense of touch consists of both kinesthetic and cutaneous (tactile) sensing [1]. Therefore, the fusion of the two sensing modalities could be beneficial for the object recognition tasks. In [21], the proprioceptive data (finger configurations/positions) and tactile features for a whole palpation sequence were concatenated into one description for object classification; the information of tactile features and contact points is combined but the information of the positions where specific tactile features were collected was lost. In recent work [22], an underactuated robot hand, with a row of TakkTile tactile sensors embedded in each link of robot fingers, was employed for object classification. The actuator positions and force sensor values form the feature space to classify object classes using random forests but there were no exploratory motions involved, with data acquired during a single and unplanned grasp.

## III. ITERATIVE CLOSEST LABELED POINT

As previously mentioned, at each collecting data point, both the tactile reading and the 3D location of the tactile sensor are recorded simultaneously. The proposed Iterative Closest Labeled Point (iCLAP) algorithm combines both appearance features obtained from tactile images and spatial distributions of objects in space. The iCLAP algorithm is introduced in detail in this section and it consists of two steps, i.e., feature label creation and iterative search for closest 4D point cloud model.

### A. Feature label creation

To begin with, a codebook of tactile features is formed from the training tactile readings and each tactile feature is assigned a label by "indexing" the codebook. As shown in Fig. 2, in the training phase a codebook/dictionary is formed by clustering the feature descriptors extracted from the tactile images using $k$-means clustering. Here, $k$ is the dictionary size, i.e., the number of clusters. The descriptors of training/test objects are then assigned to the nearest codewords/clusters in Euclidean distance and labeled with the codeword numbers $p_w = [1, 2, \cdots, k]$.

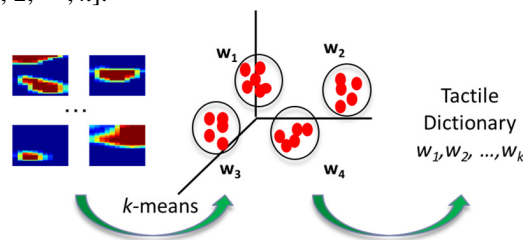

Fig. 2. The creation of the dictionary of tactile features by $k$-means clustering.

### B. Iterative search for closest 4D point cloud model

With the feature labels created from tactile readings and the movements of tactile sensors in 3D space, the object can be represented in a four-dimensional space. A single 4D point is represented by a tuple ($p_x$, $p_y$, $p_z$, $p_w$), in which $p_x$, $p_y$, $p_z$ and $p_w$ are the *x, y, z* coordinates in 3D space of tactile sensor and word label assigned to this location respectively. To calculate the mutual distance between 4D sparse data points $P$ in the test set to the model (reference) point clouds $Q$ in the training set, we extend the Iterative Closest Point (ICP) algorithm [23] to 4D space. The inputs of the iterative algorithm are the 4D model and test data point clouds, initial estimation of the transformation to align the test point cloud to the model point cloud, criteria for terminating the iterations. It can be divided into following iterative steps.

1. Let data point $p_i=\{p_{ix}, p_{iy}, p_{iz}, p_{iw}\}$ and model data point $q_i=\{q_{ix}, q_{iy}, q_{iz}, q_{iw}\}$ denote an associated set of the $N$ matched point pairs. With the 4×4 rotation matrix $R$ and 4×1 translation vector $\vec{T}$, $p_i$ can be transformed into the model point cloud's coordinate system:

$$\bar{p}_i = Rp_i + \vec{T}. \quad (1)$$

2. The second step is to find the nearest point in the model point cloud in 4D space for each transformed test data point $\bar{p}_i$. To speed up this process, a k-d tree of the model point cloud is constructed [24], [25].

3. An error metric $E_{iCLAP}$ is defined to evaluate the mean square root distance of the associated point pairs and it is minimized with the optimal rigid rotation matrix $R$ and translation vector $\vec{T}$. (see definition of $E_{iCLAP}$ in section III.C)

4. The iCLAP is iterated until any of the termination conditions is reached: Error metric $E_{iCLAP}$ > preset tolerance; Number of iterations > preset maximum number of iterations $n_{max}$; the relative change in the error metrics of two consecutive iterations falls below a predefined threshold.

### C. Error metric $E_{ICLaP}$

The error metric $E_{ICLaP}$ is defined as:

$$E_{iCLAP}(P,Q) = \sum_{i=1}^{N} \|Rp_i + \vec{T} - q_i\|^2$$

$$= \sum_{i=1}^{N} \sqrt{\begin{aligned}(\overline{p_{ix}} - q_{ix})^2 + (\overline{p_{iy}} - q_{iy})^2 + \\ (\overline{p_{iz}} - q_{iz})^2 + (\overline{p_{iw}} - q_{iw})^2\end{aligned}} \quad (2)$$

where $R$ and $\vec{T}$ are the 4×4 rotation matrix and 4×1 translation vector respectively, and data point $p_i=\{p_{ix}, p_{iy}, p_{iz}, p_{iw}\}$ and model data point $q_i=\{q_{ix}, q_{iy}, q_{iz}, q_{iw}\}$ are an associated set.

The closed form solution for minimization of $E_{ICLaP}$ is derived as following.

The centroids of $P$ and $Q$ are defined as

$$\bar{p} = \frac{1}{n}\sum_{i=1}^{n} p_i \qquad \bar{q} = \frac{1}{m}\sum_{i=1}^{m} q_i \quad (3)$$

where $n$ and $m$ are the number of test data points and model data points respectively. In general case, $n=N$. Thus the point deviations from the centroids are obtained as

$$p'_i = p_i - \bar{p} \qquad q'_i = q_i - \bar{q} \quad (4)$$

The error metric can be rewritten as

$$E_{iCLAP}(P,Q) = \sum_{i=1}^{N} \|R(p'_i + \bar{p}) + \vec{T} - (q'_i + \bar{q})\|^2 \quad (5)$$

$$= \sum_{i=1}^{N} \|Rp'_i - q'_i + (R\bar{p} - \bar{q} + \vec{T})\|^2$$

In order to minimize the error metric, the translation vector $\vec{T}$ is chosen to move the rotated data centroid to the model centroid

$$\vec{T} = \bar{q} - R\bar{p}. \quad (6)$$

Therefore, the error metric is simplified as

$$E_{iCLAP}(P,Q) = \sum_{i=1}^{N} \|Rp'_i - q'_i\|^2$$

$$= RR^T \sum_{i=1}^{N} \|p'_i\|^2 - 2tr(R\sum_{i=1}^{N} p'_i q'^T_i) + \sum_{i=1}^{N} \|q'_i\|^2$$

$$= \sum_{i=1}^{N} \|p'_i\|^2 - 2tr(R\sum_{i=1}^{N} p'_i q'^T_i) + \sum_{i=1}^{N} \|q'_i\|^2 \quad (7)$$

Now let $H = \sum_{i=1}^{N} p'_i q'^T_i$. In an expanded form, we have

$$H = \begin{bmatrix} M_{xx} & M_{xy} & M_{xz} & M_{xw} \\ M_{yx} & M_{yy} & M_{yz} & M_{yw} \\ M_{zx} & M_{zy} & M_{zz} & M_{zw} \\ M_{wx} & M_{wy} & M_{wz} & M_{ww} \end{bmatrix},$$

where $M_{ab} = \sum_{i=1}^{N} p'_{ia} q'_{ib}$, $a, b \in \{x, y, z, w\}$. To minimize the error metric $E_{ICLaP}$ the trace $tr(RH)$ has to be maximized. Let the columns of $H$ and the rows of $R$ be denoted $h_j$ and $r_j$ respectively, where $j \in \{1, 2, 3, 4\}$. The trace of $RH$ can be expanded as

$$tr(RH) = \sum_{j=1}^{4} r_j \cdot h_j \leq \sum_{j=1}^{4} \|r_j\| \|h_j\| \quad (8)$$

where the inequality is just a reformulation of the Cauchy – Schwarz inequality. Since the rotation matrix $R$ is orthogonal by definition for orthogonal transformation, its row vectors all have unit length. This implies

$$tr(RH) \leq \sum_{j=1}^{4} \sqrt{h_j^T h_j} = tr(\sqrt{H^T H}) \quad (9)$$

where the square root is taken in the operator sense. Consider the singular value decomposition of $H = U\Sigma V^T$. By choosing the rotation vector as $R = VU^T$, the trace of $RH$ becomes

$$\begin{aligned} tr(VU^T U \Sigma V^T) &= tr(V \Sigma V^{-1}) \\ &= tr(\sqrt{V\Sigma^T \Sigma V^{-1}}) \\ &= tr(\sqrt{H^T H}) \end{aligned} \quad (10)$$

which according to Eq. (9) is as large as possible. In this way, the error metric $E_{iCLAP}$ is minimized with found optimal rotation matrix $R = VU^T$ and translation vector $\vec{T} = \bar{q} - R\bar{p}$.

The obtained distances between the test point cloud and the reference models in the training set are then normalized to the sum of their squares. In this way, the distances between the test point cloud and the reference models in the training set are obtained. By comparing the error metric $E_{iCLAP}$ a model with the minimum $E_{iCLAP}$ can be found and its identity is assigned to the test object.

## IV. DATA ACQUISITION SYSTEM

The experimental setup, illustrated in Fig. 3, consists of two parts, i.e., the tactile sensor and the positioning device.

### A. WTS tactile sensor

A tactile sensor WTS 0614-34A from Weiss Robotics is used in our experiments. It has an array of 14×6 sensing elements with a size of 51mm × 24mm and a spatial resolution of 3.4mm × 3.4mm for each element. The sensing array is covered by elastic rubber foam to conduct the exerted force. As the sensor interacts with objects, the foam gets compressed and the force is transferred to the sensor; thus the pressure values change. The maximum scanning rate of the sensor is 270 frames/s but a rate of 5 frames/s was used in our experiments because in initial studies the chosen sampling rate was found to be sufficient for the tasks [3], [17].

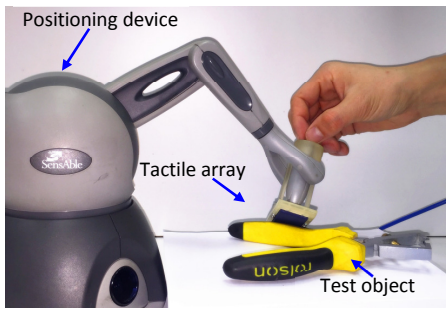

Fig. 3. The experimental setup comprises a Phantom Omni device as a robotic arm and a tactile sensor attached to its stylus.

### B. Positioning device

A Phantom Omni device with six degrees of freedom (DoF) from Sensable Technologies was used for positioning the tactile sensor and its stylus served as a robotic manipulator. It is based on a serial architecture that the stylus is linked to the housing by a single serial chain. The position of the end-effector of the stylus can be obtained and it has a nominal position resolution of around 0.055 mm. As the tactile sensor is attached to the stylus and its surface center is aligned with the end point of the stylus, the position of end-effector can be taken as the position of tactile sensor in the world coordinates.

### C. Data collection

During the data acquisition, each object was explored five times. Each exploration procedure was initialized with an idle load, namely, no object-sensor interaction. As in [11], the stylus was controlled manually to explore the object while keeping sensor plane normal to the object surface; in this manner, the object surface was covered while a number of tactile observations and sensor movement data were collected. As a result, 8492 tactile readings with corresponding contact locations for 20 objects were collected. The objects in the experiments were taken from either the lab environment or daily life with two exceptions, i.e., 3D printed point array on a flatbed and character E on a hemisphere. All of the objects are illustrated and labeled in Fig. 4. It can be noticed that there are some objects with similar appearance or spatial distributions. For instance, the sizes of the plier 1 and plier 2 are quite close and they have a similar frame, but they have different local appearance, i.e., the shape of jaws. On the other hand, some objects have similar appearance but have different spatial distributions. For example, fixed wrenches 1 and 2 have similar local appearance, but they have different sizes and spatial distributions. These similarities will test the robustness of different algorithms. Fig. 5 shows the sampled tactile images of 7 example objects. As seen in the tactile images, prominent features can be observed for each object.

## V. RESULTS AND ANALYSIS

A leave-one-out cross validation was carried out by using different dataset as test data to validate the results; the recognition rates shown as follows are the averages of cross validation results. The general goal is to achieve high recognition rate while minimize the needed amount of samples.

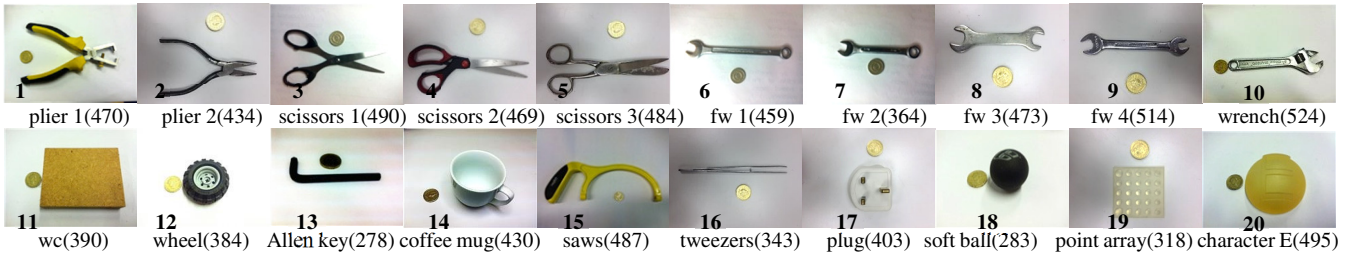

Fig. 4. Objects used for the experiments and they are labeled from 1 to 20 marked at the bottom left of the picture of each object. The name and number of collected tactile readings are also listed under each object picture. Notation: fw and wc stand for fixed wrench and wooden cuboid respectively.

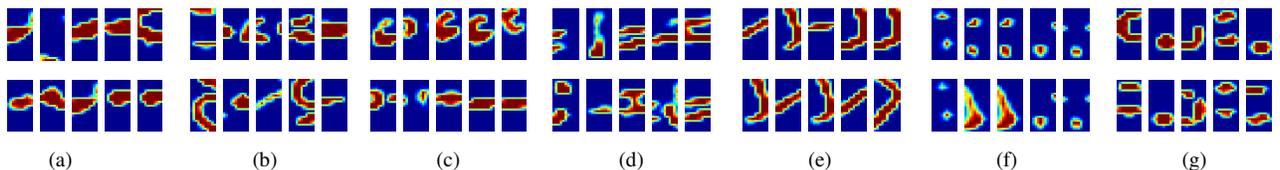

Fig. 5. Sampled tactile readings of (a) plier 1 (b) scissors 1 (c) fixed wrench 2 (d) wrench (e) Allen key (f) plug (g) 3D printed character E on a hemisphere. In these readings, prominent features can be observed for each object. The tactile images are interpolated for visualization whereas raw data were utilized in the processing.

## A. Recognition performances of BoW framework with different tactile features

To select the tactile features, we first compared the recognition performances with tactile information only (BoW framework). In total, four different features are used and compared, i.e., the Tactile-SIFT descriptors proposed in our previous work [17] and three previous features in the literature, i.e., Zernike moments (the best performing feature used in [15]), normalized Hu's moments [22], raw image moments (up to order 2) [23]. Based on [17], the dictionary size $k$ was set to 50 through the experiments. As shown in Fig. 6, the recognition performances of all the methods improve as the number of touches increase. On the other hand, our Tactile-SIFT descriptors and Zernike moments clearly outperform the other two descriptors through the experiments. Therefore, Tactile-SIFT descriptors and the Zernike moments are utilized in the following experiments.

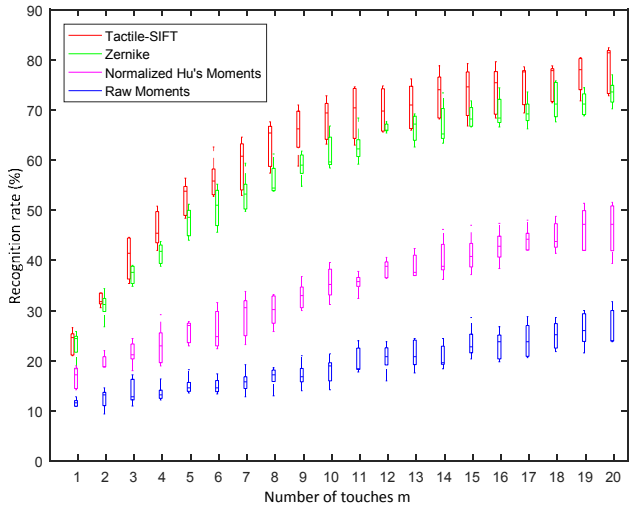

Fig. 6. The recognition rates with tactile information only (*BoW*) against different number of touches, using our Tactile-SIFT descriptors [17], Zernike moments [15], normalized Hu's moments [22] and raw image moments [23] respectively.

## B. Recognition performances of iCLAP vs methods using single sensing modalities

By using Tactile-SIFT descriptors as the tactile features, the recognition performances of BoW framework (tactile information only), ICP (kinesthetic cues only) and iCLAP algorithm against different number of object-sensor contacts, from 1 to 20, are illustrated in Fig. 7. In general, as the number of contacts increases, all of the performances of three approaches are improved. The classification results can be divided into three stages. When the tactile sensor contacts the test object for just a few times, i.e., in the case of less than 3 touches, the tactile sensing can give better view of the object than the kinesthetic cues, since tactile images are more likely to capture "first glance" appearance features of the object. On the other side, our iCLAP algorithm is 14.76% more accurate than using only kinesthetic cues, while performing similarly to tactile images. As the number of contacts increase, the recognition rates of our proposed iCLAP algorithm increment dramatically and it performs much better than those with only one modality, showing that combining the information from two sensing modalities outperforms those using only one sensing modality. When the number of contacts is greater than 12, the recognition rates of all the three methods grow slightly but our proposed iCLAP algorithm still outperform the other two methods with single sensing modalities. With 20 touches, iCLAP can achieve an average recognition rate of 80.28%.

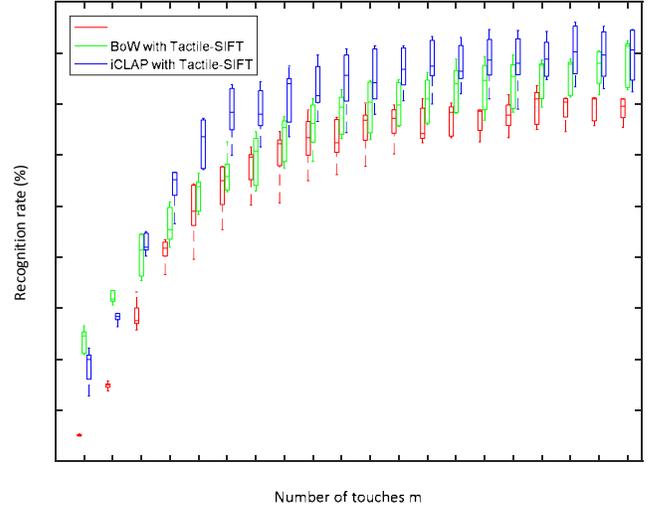

Fig. 7. The recognition rates with kinesthetic cues only (*ICP*), tactile information only (*BoW*) and dual sensing modalities using our proposed iCLAP algorithm (*iCLAP*) against different number of touches, using Tactile-SITT descriptors as tactile features.

We also compared the recognition performances of the three methods (ICP, BoW and iCLAP) by using Zernike moments as the tactile features, illustrated in Fig. 8. The recognition accuracy follows a similar pattern to the case using Tactile-SIFT descriptors. And it can be observed that the proposed iCLAP algorithm outperforms the methods using single modalities by up to 18.00% and can achieve a high average recognition rate of 85.36% with 20 touches.

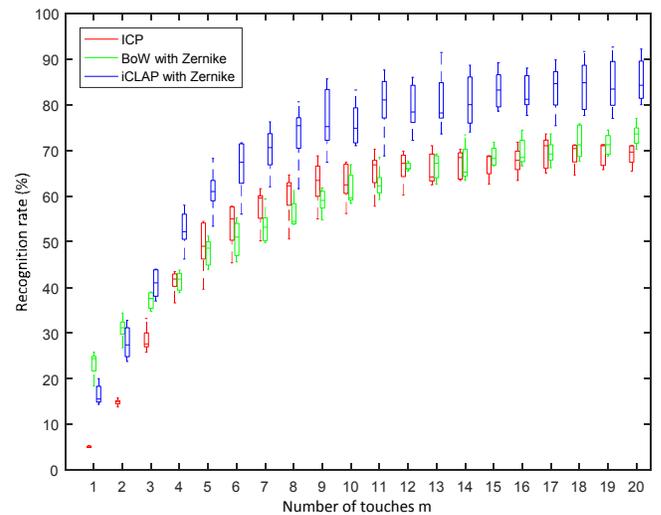

Fig. 8. The recognition rates with kinesthetic cues only (*ICP*), tactile information only (*BoW*) and dual sensing modalities using our proposed iCLAP algorithm (*iCLAP*) against different number of touches, using Zernike moments as tactile features.

In Table 1, the recognition performance improvements of iCLAP are listed. Based on the observations of the recognition performances, it is reasonable to arrive at a conclusion that our proposed iCLAP algorithm benefits from both tactile and kinesthetic sensing channels and achieves a better perception of the interacted objects.

TABLE I. Recognition performance improvement using different descritpors

| Descriptors | BoW | iCLAP | Improvement |
|---|---|---|---|
| Tactile-SIFT | 78.32% | 80.28% | 14.76% |
| Zernike moment | 73.40% | 85.36% | 18.00% |
| Hu's moment | 46.40% | ------ | ------ |
| Raw moment | 26.60% | ------ | ------ |

Note: all the BoW and iCLAP recognition rates listed in the table are the best performance in each case; the improvements are the largest differences between the recognition rates of iCLAP and BoW or ICP observed throughout the experiments ($m$ from 1 to 20).

## VI. CONCLUSION AND DISCUSSION

The experimental results of classifying 20 real objects show that iCLAP can improve the recognition performance largely compared to the methods using only one sensing channel. The proposed iCLAP algorithm can be applied to several other fields, e.g., computer vision related applications. For instance, in scene classification, as the landscape observations are correlated with the locations in which they are collected, the proposed iCLAP combining the two sensing modalities is expected to enhance the classification performance.

There are several directions to extend our work. As only the word label is utilized in the iCLAP to represent the tactile data, there is information loss to certain extent. Therefore, it will be studied how the word labels effects the convergence of the algorithm and it is also planned to include more clues of tactile patterns in the future designed algorithm, and compare these developed algorithms with other fusion methods. In this paper, the experimental data are manually collected. Therefore, the contact force between tactile sensors and explored objects was difficult to control. However, the force level might affect the tactile readings. To this end, it will be investigated how the contact force affects the recognition results and autonomous exploration of objects will be designed. In addition, it will also be studied to recognize objects with multiple tactile sensing pads. Furthermore, more objects of different types, e.g., the shape and size of object, soft object or moveable object, will be used to evaluate the algorithm.

## VII. ACKNOWLEDGMENTS

This work was partially supported by the King's-China Scholarship Council.